\pdfoutput=1
\documentclass[11pt]{article}

\usepackage{naacl2021}

\usepackage{times}
\usepackage{latexsym}

\usepackage[T1]{fontenc}
\usepackage[utf8]{inputenc}

\usepackage{microtype}

\usepackage{booktabs}
\usepackage{multirow}
\usepackage{graphicx}
\usepackage{pifont}
\usepackage{amsmath}
\usepackage{amssymb}
\usepackage{ifthen}
\usepackage{dsfont}

\usepackage{tikz}
\usetikzlibrary{bayesnet}
\usepackage{subfloat}
\usepackage{subfig}
\usepackage[linguistics]{forest}

\usepackage[ruled,vlined]{algorithm2e}

\newcommand{\prob}[2][]{\text{\bf P}\ifthenelse{\not\equal{}{#1}}{_{#1}}{}\!\left(#2\right)}
\newcommand{\expect}[2][]{\text{\bf E}\ifthenelse{\not\equal{}{#1}}{_{#1}}{}\!\left[#2\right]}
\newcommand{\var}[2][]{\text{\bf Var}\ifthenelse{\not\equal{}{#1}}{_{#1}}{}\!\left[#2\right]}

\DeclareMathOperator{\argmax}{argmax}

\newcommand{\bx}{\mathbf{x}}

\newcommand{\by}{\mathbf{y}}

\newcommand{\swap}{\textsc{swap}}
\newcommand{\reduce}{\textsc{reduce}}
\newcommand{\shift}{\textsc{shift}}
\newcommand{\merge}{\textsc{merge}}
\newcommand{\pred}[1]{\textsc{pred(#1)}}
\newcommand{\subgraph}[1]{\textsc{subgraph(#1)}}

\newcommand{\lab}{\textsc{label}}
\newcommand{\cp}{\textsc{copy}}

\newcommand{\la}[1]{\textsc{la(#1)}}
\newcommand{\ra}[1]{\textsc{ra(#1)}}
\newcommand{\id}{\textsc{id}}
\title{AMR Parsing with Action-Pointer Transformer}

\author{Jiawei Zhou$^\text{\ding{61}}$ \ \ \ Tahira Naseem$^*$ \ \ \ Ram\'{o}n Fernandez Astudillo$^*$ \ \ \ Radu Florian$^*$ \\
        $^\text{\ding{61}}$Harvard University \hspace{1cm} $^*$IBM Research \\ 
        $^\text{\ding{61}}$\texttt{\small{jzhou02@g.harvard.edu}} \ \ $^*$\texttt{\small{\{tnaseem,raduf\}@us.ibm.com}} \ \
        $^*$\texttt{\small{ramon.astudillo@ibm.com}}
        }
\begin{document}
\maketitle
\begin{abstract}
Abstract Meaning Representation parsing is a sentence-to-graph prediction task where target nodes are not explicitly aligned to sentence tokens. However, since graph nodes are semantically based on one or more sentence tokens, implicit alignments can be derived. Transition-based parsers operate over the sentence from left to right, capturing this inductive bias via alignments at the cost of limited expressiveness.
In this work, we propose a transition-based system that combines hard-attention over sentences with a target-side action pointer mechanism to decouple source tokens from node representations and address alignments.
We model the transitions as well as the pointer mechanism through straightforward modifications within a single Transformer architecture. Parser state and graph structure information are efficiently encoded using attention heads.
% Here we show that this limitation can be overcome by decoupling surface token and node representations. We represent nodes by masking one head of the self-attention mechanism of a Transformer decoder and point to node-creating actions to generate edges. We further encode updates to graph structure by repurposing edge-creating actions and relate node and surface token representations through masking of the cross-attention mechanism. 
We show that our action-pointer approach leads to increased expressiveness and attains large gains (+1.6 points) against the best transition-based AMR parser in very similar conditions. While using no graph re-categorization, our single model yields the second best \textsc{Smatch} score on AMR 2.0 (81.8), which is further improved to 83.4 with silver data and ensemble decoding.
% Our action-pointer model attains large gains (+1.6) against the best transtion-based AMR parser in very similar conditions and yields second best score overall (83.4) in terms of Smatch on AMR2.0+silver while using no graph recategorization.
\end{abstract}

\section{Introduction}

\begin{figure}
    \centering
    \includegraphics[width=\columnwidth]{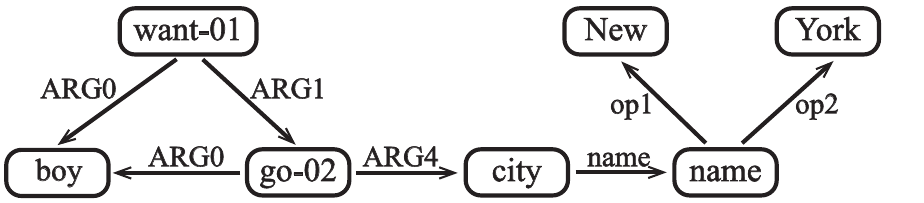}
    \caption{AMR graph expressing the meaning of the sentence \textit{The boy wants to go to New York}.}
    \label{fig:amr_example}
\end{figure}

%% what is AMR
% RAMON: Adde more explicit description of what AMR is and nodes not being aligned
Abstract Meaning Representation (AMR) \citep{banarescu2013abstract} is a sentence level semantic formalism encoding \textit{who does what to whom} in the form of a rooted directed acyclic graph. Nodes represent concepts such as entities or predicates which are not explicitly aligned to words, and edges represent relations such as subject/object (see Figure~\ref{fig:amr_example}).

AMR parsing, the task of generating the graph from a sentence, is nowadays tackled with sequence to sequence models parametrized with neural networks.
% For the purpose of this work, it is useful to distinguish two broad categories within these parsers.
There are two broad categories of methods that are highly effective in recent years.
Transition-based approaches predict a sequence of actions given the sentence. These actions generate the graph while processing tokens left-to-right through the sentence and store intermediate representations in memories such as stack and buffer \cite{wang2015transition,damonte2016incremental,ballesteros2017amr,vilares2018transition,naseem2019rewarding,astudillo2020transition,lee2020pushing}. General graph-based approaches, on the other hand, directly predict nodes and edges in sequential order from graph traversals such as breath first search or depth first search \citep{zhang2019amr,zhang2019broad,cai2019core,cai2020amr}.
While not modeling the local semantic correspondence between graph nodes and source tokens, the approaches achieve strong results without restrictions of transition-based approaches, but often require graph re-categorization, a form of graph normalization, for optimal performance.
% These approaches achieve the current state-of-the-art without the restrictions of transition-based approaches, but require graph re-categorization, a form of graph normalization, for optimal performance.

% RAMON: this should be in related work
%transition-based algorithms \citep{nivre2008algorithms} derived from dependency parsing have been often applied.
%Transition-based systems process the sentence left-to-right and generate nodes by transforming surface words. Since nodes and words are not explicitly related in AMR, various extensions have been explored to overcome this problem. These include deriving AMR parses from dependency parses \citep{wang2015transition}, introducing additional lookup tables for AMR concepts \citep{damonte2016incremental,vilares2018transition}, dynamically adding extra tokens for additional nodes \citep{guo2018better, vilares2018transition}, utilizing a $\swap$ action and collapsing nodes \citep{ballesteros2017amr, naseem2019rewarding, astudillo2020transition} or a cache data structure \citep{peng2018sequence}.

\begin{figure*}[!ht]
    \centering
    \includegraphics[scale=0.46]{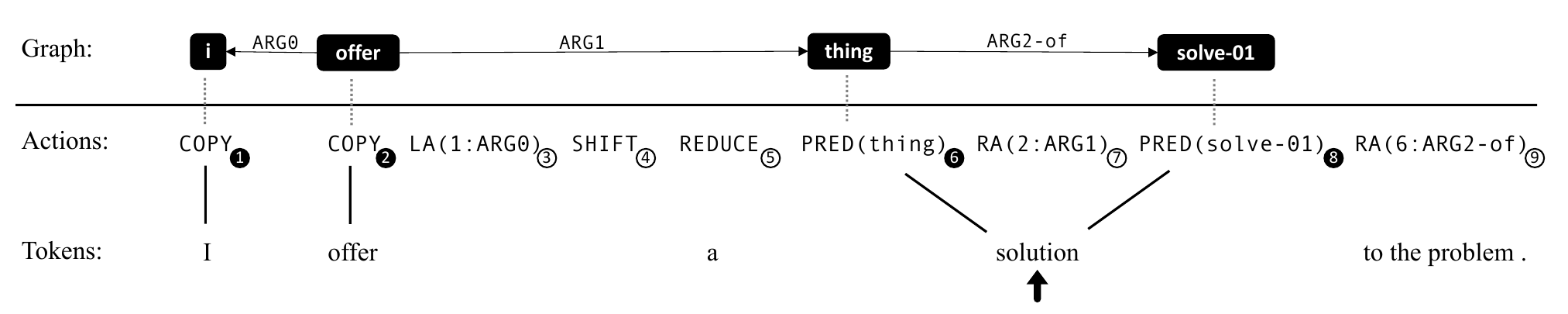}
   \caption{Source tokens, target actions and AMR graph for the sentence \textit{I offer a solution to the problem} (partially parsed). The black arrow marks the current token cursor position. The circles contain the action indices (used as ids), black circles indicate node creating actions. Only these actions are available for edge attachments. Notice that the edge actions (at steps 3, 7 and 9) explicitly refer to past nodes using the id of the action that created the node. The other participant of the edge action is implicitly assumed to be the most recently created graph node.}
    \label{fig:ex1}
\end{figure*}

%% limitation of the transition-based approaches
The strong left-to-right constraint of transition-based parsers provides a form of inductive bias that fits AMR characteristics. AMR nodes are very often normalized versions of sentence tokens and locality between words and nodes is frequently preserved. The fact that transition-based systems for AMR have alignments as the core of their explanatory model also guarantees that they produce reliable alignments at decoding time, which are useful for applications utilizing AMR parses. Despite these advantages, transition-based systems still suffer in situations when multiple nodes are best explained as aligned to one sentence token or none. Furthermore, long distance edges in AMR, e.g. re-entrancies, require excessive use of $\swap$ or equivalent actions, leading to very long action sequences. This in turn affects both a model's ability to learn and its decoding speed.

%In contrast, another view of AMR parsing treats it as a graph generation problem conditioned on the source text, loosening the local semantic correspondence between graph nodes and source tokens. The graph-based approaches \citep{zhang2019amr,zhang2019broad, cai2020amr} directly tackle node generation through seq-to-seq models and edge generation with variants of attention mechanism by comparing node representations \citep{peng2017deep, dozat2018simpler}. These approaches achieve competitive results without the restrictions of transition-based approaches, but require graph re-categorization, a form of graph normalization, for performance.

%% Our system and contribution
In this work, we propose the Action-Pointer Transition (APT) system which combines the advantages of both the transition-based approaches and more general graph-generation approaches. We focus on predicting an action sequence that can build the graph from a source sentence. The core idea is to put the target action sequence to a dual use -- as a mechanism for graph generation as well as the representation of the graph itself. Inspired by recent progress in pointer-based parsers \citep{ma-etal-2018-stack,fernandez2020transition}, we replace the stack and buffer by a cursor that moves from left to right and introduce a pointer network \citep{vinyals2015pointer} as mechanism for edge creation.  Unlike previous works, we use the pointer mechanism on the target side, pointing to past node generation actions to create edges. This eliminates the node generation and attachment restrictions of previous transition-based parsers. It is also more natural for graph generation, essentially resembling the generation process in the graph-based approaches, but keeping the graph and source aligned. 

We model both the action generation and the pointer prediction with a single Transformer model \citep{vaswani2017attention}. 
We relate target node and source token representations through masking of cross-attention mechanism, similar to \citet{astudillo2020transition} but simply with monotonic action-source alignment driven by cursor positions, rather than stack and buffer contents. Finally we also embed the AMR graph structural information in the target decoder by re-purposing edge-creating steps, and propose a novel step-wise incremental graph message passing method \citep{gilmer2017neural} enabled by the decoder self-attention mechanism.

Experiments on AMR 1.0, AMR 2.0, and AMR 3.0 benchmark datasets show the effectiveness of our APT system. We outperform the best transition-based systems %by 1.6 \textsc{Smatch} points 
while using sensibly shorter action sequences, and achieve better performance than all previous approaches with similar size of training parameters.

\section{AMR Generation with Action-Pointer}\label{sec:apt}
Figure~\ref{fig:ex1} shows a partially parsed example of a source sentence, a transition action sequence and the AMR graph for the proposed transitions. 
Given a source sentence $\bx=x_1, x_2, \ldots, x_S$, our transition system works by scanning the sentence from left to right using a cursor $c_t\in\{1, 2, \ldots, S\}$. 
Cursor movement is controlled by three actions:

\paragraph{$\shift$} moves cursor one position to the right, such that $c_{t+1}=c_t+1$. 
\paragraph{$\reduce$} is a special $\shift$ indicating that no action was performed at current cursor position.
\paragraph{$\merge$} merges tokens $x_{c_t}$ and $x_{c_t+1}$ and $\shift$s. Merged tokens act as a single token under the position of the last token merged.\\

% TODO: mention Miguel introduced merge

At cursor position $c_t$ we can generate any subgraph through following actions:

\paragraph{$\cp$} creates a node by copying the word under $x_{c_t}$. Since AMR nodes are often lemmas or propbank frames, two versions of this action exist to copy the lemma of $x_{c_t}$ or provide the first sense (frame $-01$) constructed from the lemma. This covers a large portion of the total AMR nodes. It also helps generalize for predictions of unseen nodes. We use an external lemmatizer\footnotemark\footnotetext{https://spacy.io/.} for this action.

\paragraph{$\pred{label}$} creates a node with name $\lab$ from the node names seen at train time. 

\paragraph{$\subgraph{label}$} produces an entire subgraph indexed by label $\lab$. Any future attachments can only be made to the root of the subgraph.

\paragraph{$\la{id,label}$} creates an arc with $\lab$ from last generated node to a previous node at position $\id$. Note that we can only point to past node generating actions in the action history.

\paragraph{$\ra{id,label}$} creates an arc with $\lab$ to last generated node from a previous node at position $\id$.\\

%Therefore, it is always possible to establish an edge between the last predicted node and any previous node.

Using the above actions, it is easy to derive an oracle action sequence given gold-graph information and initial word to node alignments. For current cursor position, all the nodes aligned to it are generated using $\subgraph{}$, $\cp$ or $\pred{}$ actions. Each node prediction action is followed by edge creation actions. Edges connecting to closer nodes are generated before the farther ones. When multiple connected nodes are aligned to one token, they are traversed in pre-order for node generation. A detailed description of oracle algorithm is given in Appendix \ref{a:oracle}.

\begin{figure}[!t]
    \centering
    \includegraphics[scale=0.6]{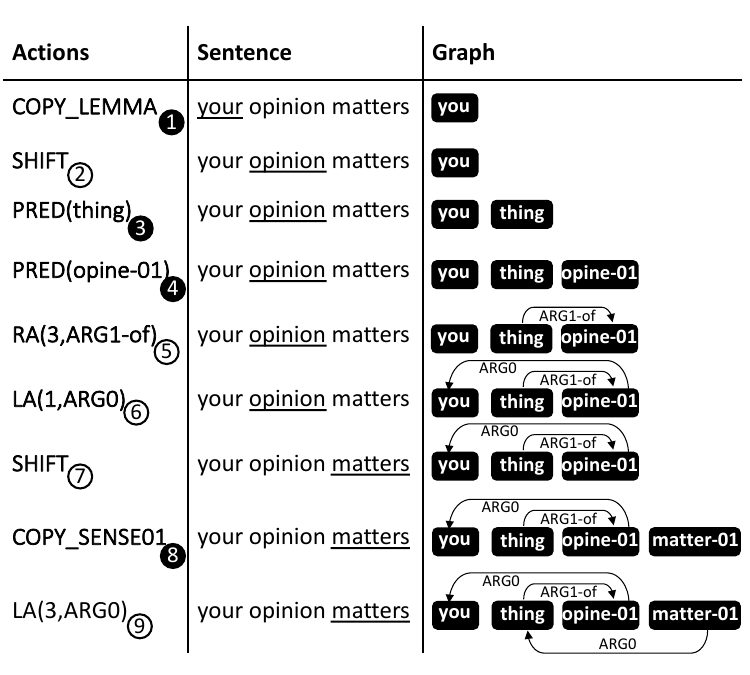}
   \caption{Step-by-step actions on the sentence \textit{your opinion matters}. Creates subgraph from a single word (\textrm{thing :ARG1-of opine-01}) and allows attachment to all its nodes. Cursor is at underlined words (post-action).}
    \label{fig:ex2}
\end{figure}

The use of a cursor variable $c_t$ decouples node reference from source tokens, allowing to produce multiple nodes and edges (see Figure \ref{fig:ex2}), even the entire AMR graph if necessary, from a single token. %\footnote{This can potentially result in an infinite action sequence, repeatedly generating nodes from same cursor position. We did not observe this behavior in trained parser at decode time. If encountered, it can be addressed by caping the maximum number of nodes allowed per cursor position.}. 
This provides more expressiveness and flexibility than previous transition-based AMR parsers, while keeping a strong inductive bias. The only restriction is that all inbound or outbound edges between current node and all previously produced nodes need to be generated before predicting a new node or shifting the cursor. This does not limit the oracle coverage, however, for trained parsers, it leads to a small percentage of disconnected graphs in decoding.  Furthermore, nodes within the $\subgraph{}$ action can not be reached for edge creation. The use of $\subgraph{}$ action, initially introduced in \citet{ballesteros2017amr}, is reduced in this work to cases where no such edges are expected, which is mainly the case for dates and named-entities. 

Compared to previous oracles \cite{ballesteros2017amr,naseem2019rewarding,astudillo2020transition}, the action-pointer does not use a $\swap$ action. It can establish an edge between the last predicted node and \emph{any} previous node, since edges are created by pointing to decoder representations. 

This oracle is expected to work with generic AMR aligners. For this work, we use the alignments generation method of \newcite{astudillo2020transition}, which generates many-to-many alignments. It is a combination of Expectation Maximization based alignments of \newcite{pourdamghani2014aligning} and rule base alignments of \newcite{flanigan2014discriminative}. Any remaining unaligned nodes are aligned based on their graph proximity to unaligned tokens. For more details, we refer the reader to the works of \newcite{astudillo2020transition} and \newcite{naseem2019rewarding}.
%The oracle assumes that in such cases, the aligned nodes make a connected subgraph which is traversed in pre-order for node generation.  %Given the more flexible mechanism, it is also guaranteed that an action sequence can be found that reproduces a given gold graph.

\section{Action-Pointer Transformer}\label{sec:model}

\subsection{Basic Architecture}

% Given a source sentence, the corresponding AMR graph can be generated following a sequence of actions based on our transition system, thus the graph generation problem is turned into a sequence generation problem, with the addition of the pointers on the target sequence itself for edge predictions. % target-side pointing
% We propose a parsing model that is based on the Transformer architecture, with a single module for all the output elements (as opposed to common graph-based approaches in \citet{zhang2019amr, cai2020amr} where different parts of graphs are learned with separate modules), and is tailored to incorporate structural information inside the parsing steps.

%We model the source to action generation with a single Transformer architecture, with an addition of the target-side self-pointing mechanism. It is also tailored to encode the transition state information and the graph structural information inside the parsing steps with decoder attentions.   

The backbone of our model is the encoder-decoder Transformer \citep{vaswani2017attention}, combined with a pointer network \citep{vinyals2015pointer}. The probability of an action sequence $\by=y_1, y_2, \ldots, y_T$ for input tokens $\bx=x_1, x_2, \ldots, x_S$ is given in our model by
\begin{equation}\label{eq:main}
    \begin{aligned}
    & \prob{\by \mid \bx} = \prod_{t=1}^T \prob{y_t \mid \mathbf{y}_{<t}, \bx} \\
    =& \prod_{t=1}^T \prob{a_t \mid \mathbf{a}_{<t}, \mathbf{p}_{<t}, \bx} \prob{p_t \mid \mathbf{a}_{\le t}, \mathbf{p}_{<t}, \bx}%\right]^{\gamma}%\nonumber
    \end{aligned}
\end{equation}
% \noindent where $a_t$, $p_t$ are action and pointer indices respectively. Pointer predictions are set to 1 when there is no pointer decision.
where at each time step $t$, we decompose the target action $y_t$ into the pointer-removed action and the pointer value with $y_t=(a_t, p_t)$. A dummy pointer $p_t=\mathrm{null}$ is fixed for non-edge actions, so that
\begin{equation*}
    \prob{p_t \mid \mathbf{a}_{\leq t}, \mathbf{p}_{<t}, \bx}=\left[\prob{p_t \mid \mathbf{a}_{<t}, \mathbf{p}_{<t}, \bx}\right]^{\gamma(a_t)}
\end{equation*}
where $\gamma(a_t)$ is an indicator variable set to 0 if $a_t$ is not an edge action and 1 otherwise.
% $\gamma$ is an indicator function that is zero unless $y_t$ is edge generating action, for which a pointer is needed. Whenever $\gamma$ is zero a dummy position target is used for $p_t$ and the indicator function prevents this operation from influencing the loss. 

Given a sequence to sequence Transformer model with $N$ encoder layers and $M$ decoder layers, each decoder layer is defined by
\begin{equation}\label{eq:decoder}
    \begin{aligned}
    \mathbf{d}^{m}_{t} = \mathrm{FF}^m(\mathrm{CA}^m(\mathrm{SA}^m(\mathbf{d}^{m-1}_{t}, \mathbf{d}^{m-1}_{\leq t}), \mathbf{e}^N))\nonumber
    \end{aligned}
\end{equation}
where $\mathrm{FF}^m()$, $\mathrm{CA}^m()$ and $\mathrm{SA}^m()$ are feed-forward, multi-head cross-attention and multi-head self-attention components respectively\footnotemark\footnotetext{Each of these are wrapped around with residual, dropout and layer normalization operations removed for simplicity.}. $\mathbf{e}^N$ is the output of last encoder layer and $\mathbf{d}^{m-1}$ is the output of the previous decoder layer, with $\mathbf{d}^0_{\leq t}$ initialized to be the embeddings of the action history $\mathbf{y}_{<t}$ concatenated with a special start symbol.
% and $\mathbf{d}^0=[\verb+<s>+; \mathbf{y}_{<t}]$ are the previous actions. 

The distribution over actions is given by
\begin{equation}
    \prob{a_t \mid \mathbf{a}_{<t}, \mathbf{p}_{<t}, \bx} = \mathrm{softmax}\left(\mathbf{W} \cdot \mathbf{d}^{M}_{t}\right)_{a_t}\nonumber
\end{equation}
\noindent where $\mathbf{W}$ are the output vocabulary embeddings, and the edge pointer distribution is given by
%
% \begin{eqnarray}
%     \prob{p_t \mid \mathbf{a}_{\le t}, \mathbf{p}_{<t}, \bx}= 
%     \nonumber\\ \mathrm{softmax}\big((\mathbf{K}_n^M \cdot \mathbf{d}^{M-1}_{<t})^T \!\!\!\!\! & \:\cdot \mathbf{Q}_n^M \cdot \mathbf{d}^{M-1}_{<t} + \mathbf{r}\big)_{p_t}\nonumber
%     \label{eq:ah2}
% \end{eqnarray}
\begin{eqnarray}
    \prob{p_t \mid \mathbf{a}_{<t}, \mathbf{p}_{<t}, \bx}= 
    \nonumber\\ \mathrm{softmax}\big((\mathbf{K}^M \cdot \mathbf{d}^{M-1}_{\leq t})^T \!\!\!\!\! & \:\cdot \mathbf{Q}^M \cdot \mathbf{d}^{M-1}_{t}\big)_{p_t}\nonumber
    \label{eq:ah2}
\end{eqnarray}
where $\mathbf{K}^M$, $\mathbf{Q}^M$ are key and query matrices of 1 head of the last decoder self-attention layer $\mathrm{SA}^M()$.
% where $\mathbf{K}_n^M$, $\mathbf{Q}_n^M$ are key and query matrices of the last decoder self-attention layer $\mathrm{SA}^M()$ and $\mathbf{r}$ a masking matrix used for the graph embedding mechanism described in the next section.
The top layer self-attention is a natural choice for the pointer network, since it is likely to have high values for the nodes involved in the edge direction and label prediction. 
% By supervising one attention head of the decoder, we also introduce a secondary task thus making actions distribution aware of the pointer distribution.
Although the edge action and its pointing value are both output at the same step, the specialized pointer head is also part of the overall self-attention mechanism used to compute the model's hidden representations, thus making actions distribution aware of the pointer distribution.

Our transition system moves the cursor $c_t$ over the source from left to right during parsing, essentially maintaining a monotonic alignment between target actions and source tokens.
We encode the alignment $c_t$ with hard attentions in cross-attention heads $\mathrm{CA}^m()$ with $m = 1 \cdots M$ at every decoder layer. We mask one head of the cross-attention to see only the aligned source token at $c_t$, and augment it with another head masked to see only positions $>c_t$. This is similar to the hard attention in \citet{peng2018sequence} and parser state encoding in \citet{astudillo2020transition}.

% Similarly to the stack-Transformer \cite{astudillo2020transition}, we also specialize cross-attention heads at every layer of the decoder $\mathrm{CA}^m()$ with $m = 1 \cdots M$ to reflect the parser state. We mask one head of the cross attention to see only the tokens with positions bigger than $c_t$ and another head to point only to position $c_t$.

As in prior works, we restrict the output space of our model to only allow valid actions given $\mathbf{x}, \mathbf{y}_{<t}$.
The restriction is not only enforced at inference, but is also internalized with the model during training so that the model can always focus on relevant action subsets when making predictions.

\begin{figure*}
    \centering
\hspace{-1cm}
\begin{tikzpicture}
\node (rect) at (7,1) [draw,rounded corners,thick,minimum width=15cm,minimum height=2.6cm] {};

% circles

%\node at (0,-1.6) [] (w0j){\footnotesize Action history:};
%\node at (-2,1) [] (w0j){\footnotesize $3$ Layer decoder:};
%\node at (0,3.2) [] (w0j){\footnotesize Target Actions:};

% warping for horizontal separation
\def \xwarp {1.4}

% 0
\def \xpos {0}
% network nodes
\node at ({\xpos * \xwarp},0) [circle,draw,thick,inner sep=2.5pt](i\xpos){};
\node at ({\xpos * \xwarp},1) [circle,draw,thick,inner sep=2.5pt](o\xpos){};
\node at ({\xpos * \xwarp},2) [circle,draw,thick,inner sep=2.5pt](a\xpos){};
% vertical line
\path (i\xpos) edge[-,thick] (o\xpos);
\path (o\xpos) edge[-,thick] (a\xpos);
% action sequences
\node at ({\xpos * \xwarp},-0.6) [] (w\xpos i){\footnotesize \verb+<s>+};
\node at (0,-1) [] (w\xpos i){\footnotesize Index:};
\node at ({\xpos * \xwarp},2.6) [] (w\xpos j){\footnotesize REDUCE};

% 1
\def \xpos {1}
% network nodes
\node at ({\xpos * \xwarp},0) [circle,draw,thick,inner sep=2.5pt](i\xpos){};
\node at ({\xpos * \xwarp},1) [circle,draw,thick,inner sep=2.5pt](o\xpos){};
\node at ({\xpos * \xwarp},2) [circle,draw,thick,inner sep=2.5pt](a\xpos){};
\node at ({\xpos * \xwarp},-1) [](index\xpos){\footnotesize {1}};
% vertical line
\path (i\xpos) edge[-,thick] (o\xpos);
\path (o\xpos) edge[-,thick] (a\xpos);
% action sequences
\node at ({\xpos * \xwarp},-0.6) [] (w\xpos i){\footnotesize REDUCE};
\node at ({\xpos * \xwarp},2.6) [] (w\xpos j){\footnotesize COPY};

% 1
\def \xpos {2}
% network nodes
\node at ({\xpos * \xwarp},0) [circle,draw,thick,inner sep=2.5pt](i\xpos){};
\node at ({\xpos * \xwarp},1) [circle,draw,thick,inner sep=2.5pt](o\xpos){};
\node at ({\xpos * \xwarp},2) [circle,draw,thick,inner sep=2.5pt](a\xpos){};
\node at ({\xpos * \xwarp},-1) [](index\xpos){\footnotesize {2}};
% vertical line
\path (i\xpos) edge[-,thick] (o\xpos);
\path (o\xpos) edge[-,thick] (a\xpos);
% action sequences
\node at ({\xpos * \xwarp},-0.6) [] (w\xpos i){\footnotesize COPY};
\node at ({\xpos * \xwarp},2.6) [] (w\xpos j){\footnotesize SHIFT};
% graph
\node [draw,rounded corners,thick] (node0) at ({\xpos * \xwarp},-1.8) {\footnotesize boy};

% 1
\def \xpos {3}
% network nodes
\node at ({\xpos * \xwarp},0) [circle,draw,thick,inner sep=2.5pt](i\xpos){};
\node at ({\xpos * \xwarp},1) [circle,draw,thick,inner sep=2.5pt](o\xpos){};
\node at ({\xpos * \xwarp},2) [circle,draw,thick,inner sep=2.5pt](a\xpos){};
\node at ({\xpos * \xwarp},-1) [](index\xpos){\footnotesize {3}};
% vertical line
\path (i\xpos) edge[-,thick] (o\xpos);
\path (o\xpos) edge[-,thick] (a\xpos);
% action sequences
\node at ({\xpos * \xwarp},-0.6) [] (w\xpos i){\footnotesize SHIFT};
\node at ({\xpos * \xwarp},2.6) [] (w\xpos j){\footnotesize COPY};
% graph
%\node [draw,rounded corners,thick] (pict1) at (\xpos,-1.8) {\footnotesize boy};

% 2
\def \xpos {4}
% network nodes
\node at ({\xpos * \xwarp},0) [circle,draw,thick,inner sep=2.5pt](i\xpos){};
\node at ({\xpos * \xwarp},1) [circle,draw,thick,inner sep=2.5pt](o\xpos){};
\node at ({\xpos * \xwarp},2) [circle,draw,thick,inner sep=2.5pt](a\xpos){};
\node at ({\xpos * \xwarp},-1) [](index\xpos){{\footnotesize 4}};
% vertical line
\path (i\xpos) edge[-,thick] (o\xpos);
\path (o\xpos) edge[-,thick] (a\xpos);
% action sequences
\node at ({\xpos * \xwarp},-0.6) [] (w\xpos i){\footnotesize COPY};
\node at ({\xpos * \xwarp},2.6) [] (w\xpos j){\footnotesize LA(2)};
% graph
\node [draw,rounded corners,thick] (node1) at ({\xpos * \xwarp},-1.8) {\footnotesize want-01};

% 3
\def \xpos {5}
% network nodes
\node at ({\xpos * \xwarp},0) [circle,draw,thick,inner sep=2.5pt](i\xpos){};
\node at ({\xpos * \xwarp},1) [circle,draw,thick,inner sep=2.5pt](o\xpos){};
\node at ({\xpos * \xwarp},2) [circle,draw,thick,inner sep=2.5pt](a\xpos){};
\node at ({\xpos * \xwarp},-1) [](index\xpos){\footnotesize {5}};
% vertical line
\path (i\xpos) edge[-,thick] (o\xpos);
\path (o\xpos) edge[-,thick] (a\xpos);
% action sequences
%\node at ({\xpos * \xwarp},-0.6) [] (w\xpos i){\footnotesize LA(2)};
\node at ({\xpos * \xwarp},-0.6) [opacity=0.7] (w\xpos i){\footnotesize COPY};
\node at ({\xpos * \xwarp},2.6) [] (w\xpos j){\footnotesize SHIFT};
% graph
\node [draw,rounded corners,thick] (node2) at ({\xpos * \xwarp},-1.8) {\footnotesize want-01};

% 4
\def \xpos {6}
% network nodes
\node at ({\xpos * \xwarp},0) [circle,draw,thick,inner sep=2.5pt](i\xpos){};
\node at ({\xpos * \xwarp},1) [circle,draw,thick,inner sep=2.5pt](o\xpos){};
\node at ({\xpos * \xwarp},2) [circle,draw,thick,inner sep=2.5pt](a\xpos){};
\node at ({\xpos * \xwarp},-1) [](index\xpos){\footnotesize {6}};
% vertical line
\path (i\xpos) edge[-,thick] (o\xpos);
\path (o\xpos) edge[-,thick] (a\xpos);
% action sequences
\node at ({\xpos * \xwarp},-0.6) [] (w\xpos i){\footnotesize SHIFT};
\node at ({\xpos * \xwarp},2.6) [] (w\xpos j){\footnotesize REDUCE};

% 5
\def \xpos {7}
% network nodes
\node at ({\xpos * \xwarp},0) [circle,draw,thick,inner sep=2.5pt](i\xpos){};
\node at ({\xpos * \xwarp},1) [circle,draw,thick,inner sep=2.5pt](o\xpos){};
\node at ({\xpos * \xwarp},2) [circle,draw,thick,inner sep=2.5pt](a\xpos){};
\node at ({\xpos * \xwarp},-1) [](index\xpos){{\footnotesize 7}};
% vertical line
\path (i\xpos) edge[-,thick] (o\xpos);
\path (o\xpos) edge[-,thick] (a\xpos);
% action sequences
\node at ({\xpos * \xwarp},-0.6) [] (w\xpos i){\footnotesize REDUCE};
\node at ({\xpos * \xwarp},2.6) [] (w\xpos j){\footnotesize PRED};

% 5
\def \xpos {8}
% network nodes
\node at ({\xpos * \xwarp},0) [circle,draw,thick,inner sep=2.5pt](i\xpos){};
\node at ({\xpos * \xwarp},1) [circle,draw,thick,inner sep=2.5pt](o\xpos){};
\node at ({\xpos * \xwarp},2) [circle,draw,thick,inner sep=2.5pt](a\xpos){};
\node at ({\xpos * \xwarp},-1) [](index\xpos){\footnotesize {8}};
% vertical line
\path (i\xpos) edge[-,thick] (o\xpos);
\path (o\xpos) edge[-,thick] (a\xpos);
% action sequences
\node at ({\xpos * \xwarp},2.6) [] (w\xpos i){\footnotesize RA(5)};
\node at ({\xpos * \xwarp},-0.6) [] (w\xpos j){{\footnotesize PRED}};
% graph
\node [draw,rounded corners,thick] (node3) at ({\xpos * \xwarp},-1.8) {\footnotesize go-02};

% 5
\def \xpos {9}
% network nodes
\node at ({\xpos * \xwarp},0) [circle,draw,thick,inner sep=2.5pt](i\xpos){};
\node at ({\xpos * \xwarp},1) [circle,draw,thick,inner sep=2.5pt](o\xpos){};
\node at ({\xpos * \xwarp},2) [circle,draw,thick,inner sep=2.5pt](a\xpos){};
\node at ({\xpos * \xwarp},-1) [](index\xpos){{\footnotesize 9}};
% vertical line
\path (i\xpos) edge[-,thick] (o\xpos);
\path (o\xpos) edge[-,thick] (a\xpos);
% action sequences
\node at ({\xpos * \xwarp},2.6) [] (w\xpos j){\footnotesize LA(2)};
%\node at ({\xpos * \xwarp},-0.6) [] (w\xpos i){\footnotesize RA(5)};
\node at ({\xpos * \xwarp},-0.6) [opacity=0.7] (w\xpos i){\footnotesize PRED};
% graph
\node [draw,rounded corners,thick] (node4) at ({\xpos * \xwarp},-1.8) {\footnotesize go-02};

% 5
\def \xpos {10}
% network nodes
\node at ({\xpos * \xwarp},0) [circle,draw,thick,inner sep=2.5pt](i\xpos){};
\node at ({\xpos * \xwarp},1) [circle,draw,thick,inner sep=2.5pt](o\xpos){};
\node at ({\xpos * \xwarp},2) [circle,draw,thick,inner sep=2.5pt](a\xpos){};
\node at ({\xpos * \xwarp},-1) [](index\xpos){{\footnotesize 10}};
% vertical line
\path (i\xpos) edge[-,thick] (o\xpos);
\path (o\xpos) edge[-,thick] (a\xpos);
% action sequences
\node at ({\xpos * \xwarp},2.6) [] (w\xpos j){\footnotesize SHIFT};
%\node at ({\xpos * \xwarp},-0.6) [] (w\xpos i){\footnotesize LA(2)};
\node at ({\xpos * \xwarp},-0.6) [opacity=0.7] (w\xpos i){\footnotesize PRED};
% graph
\node [draw,rounded corners,thick] (node5) at ({\xpos * \xwarp},-1.8) {\footnotesize go-02};

% diago-02nal lines
\path (i2) edge[-,draw=cyan!80!blue,thick] (o5);
\path (o2) edge[-,draw=cyan!80!blue,thick] (a5);
\path (i2) edge[-,draw=magenta!90!blue,thick] (o10);
\path (o2) edge[-,draw=magenta!90!blue,thick] (a10);

\path (i5) edge[-,draw,dashed,thick] (o10);
\path (o5) edge[-,draw,dashed,thick] (a10);
\path (i5) edge[-,draw=cyan!80!blue,thick] (o9);
\path (o5) edge[-,draw=cyan!80!blue,thick] (a9);

% arcs for [want-01]
%\draw [->,draw=cyan!80!blue,thick] (node2) to[bend left=16] (node0);
%\draw [->] (node2) to[bend right=16] (node4);
%\draw [->] (node2) to[bend right=20] (node5);
%\draw [->,draw=magenta!90!blue,thick] (node5) to[bend right=11] (node0);

\draw [-stealth,draw=cyan!80!blue,thick,rounded corners=3] ($(node2)-(0.2,0.2)$) --  ($(node2)-(0.2,0.5) $) -- ( $(node0)-(0,0.5)$) --  (node0);
\draw [-stealth,draw=cyan!80!blue,thick,rounded corners=3] ($ (node2) + (0.2,-0.2) $)  --  ($ (node2)+(0.2,-0.5) $) -- ( $(node4)-(0,0.5)$) --  (node4);
\draw [-stealth,dashed,rounded corners=3] (node2)  --  ($ (node2)-(0,0.7) $) -- ( $(node5)-(0,0.7)$) --   (node5);
\draw [-stealth] [draw=magenta!90!blue,thick,rounded corners=3] (node5) --  ($ (node5) + (0,0.5) $) -- ( $(node0)+(0,0.5)$) -- (node0);

\vspace{1cm}

\end{tikzpicture}

%\vspace{-0.2cm}
\caption{Encoding graph with $2$ decoder layers for the sentence \textit{The boy wants to go}. From top to bottom: target output action sequence, masked decoder self-attention, input action history and partial graph. Edge-creating action steps in the action history are used to hold updated node representations. Action labels and edge direction treatment are removed for clarity.}
    \label{fig:graph_embedding}
\end{figure*}
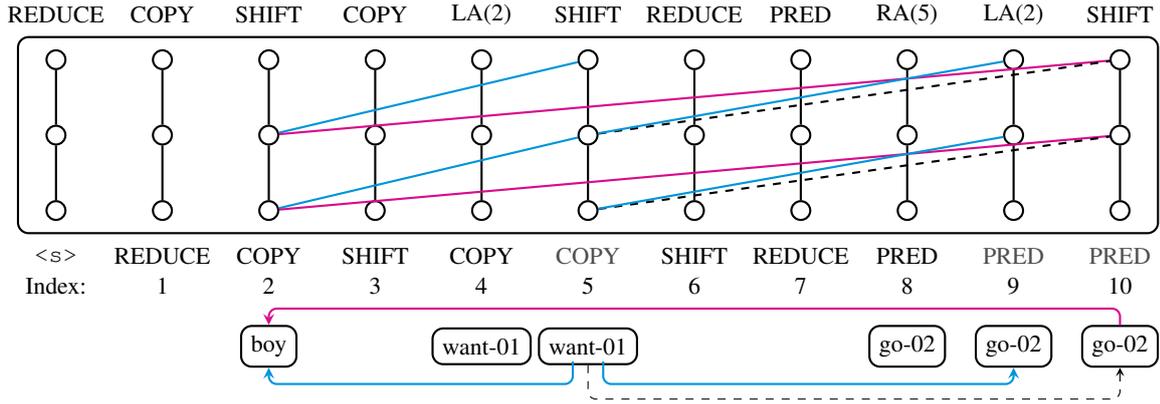

\subsection{Incremental Graph Embedding}

Incrementally generated graphs are usually modeled via graph neural networks \citep{li2018learning}, where a node's representation is updated from the collection of it's neighboring nodes' representations by message passing \citep{gilmer2017neural}.
However, this requires re-computation of all node representations every time the graph is modified, which is expensive, prohibiting its use in previous graph-based AMR parsing works \citep{cai2020amr}.
To better utilize the intermediate topological graph information without losing the efficient parallelization of Transformer, we propose to use the edge creation actions as updated views of each node, that encode this node's neighboring subgraph. This does not change the past computations and can be done by altering the hard masking of the self-attention heads of decoder layers $\mathrm{SA}^m()$ . By interpreting the decoder layers as implementing message passing vertically, we can fully encode graphs up to depth $M$.

%We update the representation of the same node at each step of the sub-sequence in a message passing fashion by restricting the decoder self-attention of each layer $\mathrm{SA}^m()$ to only attend to neighboring nodes built up to that step. Consequently, 

% We view the target actions that generate nodes as representations of corresponding graph nodes, thus 
% In the following, we use \textit{node} to refer to both a graph node and the target action that generates the node interchangeably.

%Conceptually, we use target actions that generate nodes to represent the graph nodes directly.
Given a node generating action $a_t=v$, it is followed by $k\geq0$ edge generating actions $a_{t+1}, a_{t+2}, \dots, a_{t+k}$ that connect the current node with previous nodes, pointed by $p_{t+1}, p_{t+2}, \dots, p_{t+k}$ positions on the target side.
This also defines $k$ graph modifications, expanding the graph neighborhood on the current node.
Figure~\ref{fig:graph_embedding} shows an example for the sentence \textit{The boy wants to go}, with node prediction actions at positions $t=2,4,8$, with $k$ being 0, 1, 2, respectively. We use the steps from $t$ to $t+k$ in the Transformer decoder to encode this expanding neighborhood. In particular, we fix the decoder input as the current node action $v$ for these steps, as illustrated in the input actions in Figure~\ref{fig:graph_embedding}. At each intermediate step $\tau\in[t, t+k]$, 2 decoder self-attention heads $\mathrm{SA}^m()$ are restricted to only attend to the direct graph neighbors of the current node, represented by previous nodes at positions $p_{t}, p_{t+1}, \cdots, p_{\tau}$ as well as the current position $\tau$.
This essentially builds sub-sequences of node representations with richer graph information step by step, and we use the last reference of the same node for pointing positions when generating new edges.
Moreover, when propagating this masking pattern along $m$ layers, each node encodes its $m$-hop neighborhood information.
% providing a higher level structural view of the currently built graph.
This defines a message passing procedure as shown in Figure~\ref{fig:graph_embedding}, encoding the compositional relations between nodes.
% or move to the experimental setup section
Since the edges have directions indicated by \textsc{la} and \textsc{ra}, we also encode the direction information by separating the two heads with each only considering one direction.
\section{Training and Inference}
% training and inference, pre and post processing
% including the alignment, recatogrization difference, with external NER

% \subsection{Training and Inference}
% detailed inference algorithm (constraint beam search) in the appendix

Our model is trained by maximizing the log likelihood of Equation~(\ref{eq:main}). The valid action space, action-source alignment $c_t$, and the graph embedding mask at each step $t$ are pre-calculated at training time. For inference, we modify the beam search algorithm to jointly search for actions and edge pointers and combine them to find the action sequence that maximizes Equation~(\ref{eq:main}). We also consider hard constraints in the searching process such as valid output actions and valid target pointing values at different steps to ensure an AMR graph is recoverable. For the structural information that is extracted from the parsing state such as $c_t$ and graph embedding masks, we compute them on the fly at each new step of decoding based on the current results, which are then used by the model for the next step decoding. We detail our search algorithm in Appendix \ref{b:decoding}.
% complexity?

% \subsection{Pre and Post Processing}
% talk about recategorization

% \section{Inference for AMR Parsing}
% % AMR graph generation
% % Decoding/Inference
% % use beam search with various constraints
% % post-processing: ADDNODE for a subgraph, normalization, wikification

\section{Experimental Setup}

\paragraph{Data and Evaluation}
% data
We test our approach on two widely used AMR parsing benchmark datasets: AMR 2.0 (LDC2017T10) and AMR 1.0 (LDC2014T12). The AMR graphs are all human annotated.
% The AMR 2.0 sembank task has been extensively used in recent works, encompassing named entity recognition, word sense disambiguation and co-reference resolution among other sub-tasks, consisting of 36521 sentence and AMR pairs for training.
The two datasets have 36521 and 10312 training AMRs, respectively, and share 1368 development AMRs and 1371 testing AMRs\footnote{Although there are annotation revisions from AMR 1.0 to AMR 2.0. Link to data: https://amr.isi.edu/download.html.}.
We also report results on the latest AMR 3.0 (LDC2020T02) dataset, which is larger in size but has not been fully explored, with 55635 training AMRs and 1722 and 1898 AMRs for development and testing set.
% AMR 2.0 consists of 36521 sentence and AMR pairs for training.
% AMR 1.0 shares the same development (1368 AMRs) and test (1371 AMRs) set with AMR 2.0\footnote{Although there are annotation revisions from AMR 1.0 to AMR 2.0}, but has more limited training data of 10312 AMRs, which can be used to test model's efficiency on small data.
Wiki links are removed in the pre-processing of data, and we run a wikification approach in post-processing to recover Wikipedia entries in the AMR graphs as in \citet{naseem2019rewarding}.
% (\note{JK aligner was mentioned in the transition system last section, as part of the oracle})

% alignment and oracle
% We use the JAMR and Kevin (\note{citation?}) alignments from \citet{naseem2019rewarding}, based on which we also tried force-aligning the unaligned nodes to unaligned words in order (\note{check this, and name the two alignments for later comparisons?}) to improve the oracle action sequences.
% Note that the alignments are only used to generate oracle action sequences for training, whereas at decoding time they can be generated along with the AMR graphs by our model.
% Wiki links are removed in the pre-processing of data, and we run a wikification approach in post-processing to recover Wikipedia entries in the AMR graphs as in \citet{naseem2019rewarding}.

% evaluation
For evaluation, we use the \textsc{Smatch} (F1) scores\footnote{There are small variations of \textsc{Smatch} computation due to the stochastic nature of graph matching algorithm.} \citep{cai2013smatch} and further the fine-grained evaluation metrics \citep{damonte2016incremental} to assess the model's AMR parsing performance.

\paragraph{Model Configuration}

% We implement our model taking as reference the available stack-Transformer implementation in the \textsc{fairseq} toolkit \citep{ott2019fairseq}\footnotemark\footnotetext{\url{https://github.com/IBM/transition-amr-parser}} and use the same
Our \textit{base} setup has 6 layers and 4 attention heads for both the Transformer encoder and decoder, with model size 256 and feedforward size 512.
We also compare with a \textit{small} model with 3 layers in encoder and decoder but identical otherwise.
The pointer network is always tied with one target self-attention head of the top decoder layer.
% which is supervised during training and used for decoding during testing.
We use the cross-attention of all decoder layers for action-source alignment.
For graph embedding, we use 2 heads of the bottom 3 layers for the base model and bottom 2 layers for the small model.
We use contextualized embeddings extracted from the pre-trained RoBERTa \citep{liu2019roberta} large model for the source sentence, with average of all layer states and BPE tokens mapped to words by averaging as in \cite{lee2020pushing}. The pre-trained embeddings are fixed. 
For target actions we train our own embeddings along with the model. 

\paragraph{Implementation Details}

We use the Adam optimizer with $\beta_1$ of 0.9 and $\beta_2$ of 0.98 for training. Each data batch has 3584 maximum number of tokens, and the learning rate schedule is the same as \citet{vaswani2017attention}, where we use the maximum learning rate of $5\mathrm{e}{-4}$ with 4000 warm-up steps. We use a dropout rate of 0.3 and label smoothing rate of 0.01.
% The training loss is composed of the cross-entropy of the action labels and that of the pointer values if the action corresponds to an edge, with a 1:1 ratio.
% The output distribution for each target label is restricted over the valid actions at that position by default (\note{or move this to the model setup}), which is achieved by applying a sub-vocabulary mask at each position.
We train all the models for a maximum number of 120 epochs, and average the best 5  epoch checkpoints among the last 40 checkpoints based on the \textsc{Smatch} scores on the development data with greedy decoding.
We use a default beam size of 10 for decoding.
We implement our model\footnotemark\footnotetext{Available under \url{https://github.com/IBM/transition-amr-parser}.} 
% taking as reference the available stack-Transformer implementation 
with the \textsc{fairseq} toolkit \citep{ott2019fairseq}.
All models are trained and tested on a single Nvidia Titan RTX GPU. Training takes about 10 hours on AMR 2.0 and 3.5 hours on AMR 1.0.
%Nvidia Titan RTX/V100

% For inference, we use the constraint beam search to ensure valid generations with default beam size 10. The action labels and the pointer values are jointly decoded, where the beam scores of edges are modified by combining the scores of the most likely pointer values from the pointer network with a ratio 1:1 (\note{could change during decoding}). All the graph state information during the construction is generated on the fly within the decoding and fed into the model for the next prediction. After generation, the pointer values and the action labels are then joint to recover the AMR graph for evaluation.
% All the models are implemented with the \textsc{Fairseq} toolkit \citep{ott2019fairseq}, and are trained and tested on a single %Nvidia Titan RTX/V100
% GPU.

\section{Results and Analysis}
% data oracle smatch: show model improvements from improving the oracle

\subsection{Main Results}\label{results:main}

\begin{table}[]
    \centering
    \resizebox{\columnwidth}{!}{%
    \begin{tabular}{ccc}
    \toprule
        Transition system & Avg. \#actions & Oracle \textsc{Smatch}  \\
    \midrule
         \citet{naseem2019rewarding}$^*$ & 73.6 & 93.3 \\
         \citet{astudillo2020transition}$^*$ & 76.2 & 98.0 \\
         Ours & 41.6 & 98.9 \\
    \bottomrule
    \end{tabular}
    } % resize box
    \caption{Average number of actions and oracle \textsc{Smatch} on AMR 2.0 training data. The average source length is 18.9.
    $^*$ from author correspondence.}
    % \caption{Average number of actions and oracle \textsc{Smatch} on AMR 2.0 training data compared with previous transition systems.}
    \label{tab:oracle_smatch}
\end{table}

%The AMR oracles in transition-based systems can not fully recover all graphs. 
\paragraph{Oracle Actions}
Table~\ref{tab:oracle_smatch} compares the oracle data \textsc{Smatch} and average action sequence length on the AMR 2.0 training set among recent transition systems.
Our approach yields much shorter action sequences due to the target-side pointing mechanism. It has also the best coverage on training AMR graphs, due to the flexibility of our transitions that can capture the majority of graph components. We chose not to tackle a number of small corner cases, such as disconnected subgraphs for a token, that account for the missing oracle performance.

\begin{table}[!t]
    \centering
    \resizebox{\columnwidth}{!}{%
    \begin{tabular}{cll}
        \toprule
        Corpus & Model & \textsc{Smatch} (\%) \\
        \midrule
        \multirow{14}{*}{\begin{tabular}{@{}c@{}} AMR \\ 1.0 \end{tabular}} &
         \citet{pust2015parsing} & 67.1 \\
        & \citet{flanigan2016cmu} & 66.0 \\
        % & \citet{ballesteros2017amr} & 64.0 \\
        & \citet{wang2017getting}$^G$ & 68.1 \\
        & \citet{guo2018better}$^G$ & 68.3 \small{$\pm$0.4} \\
        & \citet{zhang2019amr}$^{B,G}$ & 70.2 \small{$\pm$0.1} \\
        & \citet{zhang2019broad}$^{B,G}$ & 71.3 \small{$\pm$0.1} \\
        & \citet{cai2020amr}$^{B}$ & 74.0 \\
        & \citet{cai2020amr}$^{B,G}$ & 75.4 \\
        & \citet{astudillo2020transition}$^*\,$$^{R}$ & \textbf{76.9 \small{$\pm$0.1}} \\
        \cmidrule(lr){2-3}
        & \citet{lee2020pushing}$^{R}$ (85K silver) & \textbf{78.2 \small{$\pm$0.1}} \\
        \cmidrule(lr){2-3}
        & APT small$^{R}$ & 78.2 / 78.2 \small{$\pm$0.0} \\
        & APT base$^{R}$ & \textbf{78.5 / 78.3 \small{$\pm$0.1}} \\
        \cmidrule(lr){2-3}
        & APT small$^{R}$ p.e. & 79.7 \\
        & APT base$^{R}$ p.e. & \textbf{79.8} \\
        \midrule
        \multirow{21}{*}{\begin{tabular}{@{}c@{}} AMR \\ 2.0 \end{tabular}} & 
         \citet{van2017neural} & 71.0 \\
        & \citet{groschwitz2018amr}$^G$ & 71.0 \\
        & \citet{lyu2018amr}$^G$ & 74.4 \small{$\pm$0.2} \\
        & \citet{cai2019core} & 73.2 \\
        & \citet{lindemann2019compositional} & 75.3 \small{$\pm$0.1} \\
        & \citet{naseem2019rewarding}$^B$ & 75.5 \\
        & \citet{zhang2019amr}$^{B,G}$ & 76.3 \small{$\pm$0.1} \\
        & \citet{zhang2019broad}$^{B,G}$ & 77.0 \small{$\pm$0.1} \\
        & \citet{cai2020amr}$^{B}$ & 78.7 \\
        & \citet{cai2020amr}$^{B,G}$ & 80.2 \\
        & \citet{astudillo2020transition}$^*\,$$^{R}$ & 80.2 \small{$\pm$0.0}\\
        & \citet{bevilacqua2021one}$^\text{\ding{61}}$ & \textbf{83.8} \\
        \cmidrule(lr){2-3}
        & \citet{xu2020improving} (4M silver) & 80.2 \\
        & \citet{lee2020pushing}$^{R}$ (85K silver) & 81.3 \small{$\pm$0.0} \\
        & \citet{bevilacqua2021one}$^\text{\ding{61}}$ (200K silver) & \textbf{84.3} \\
        \cmidrule(lr){2-3}
        & APT small$^{R}$ & 81.7 / 81.5 \small{$\pm$0.2} \\
        & APT base$^{R}$  & \textbf{81.8 / 81.7 \small{$\pm$0.1}} \\
        \cmidrule(lr){2-3}
        & APT small$^{R}$ p.e.  & 82.5 \\
        & APT base$^{R}$ p.e.  & 82.8 \\
        % & APT base$^{R}$ p.e.  & 83.2 \\  % from Ramon's model
        & APT base$^{R}$ (70K Silver)  &  82.8 / 82.6 \small{$\pm 0.1$}\\
        & APT base$^{R}$ (70K Silver) p.e.  & \textbf{83.4} \\
        \midrule
        \multirow{4}{*}{\begin{tabular}{@{}c@{}} AMR \\ 3.0 \end{tabular}} & 
         \citet{lyu2020differentiable} & 75.8 \\
         & \citet{bevilacqua2021one}$^\text{\ding{61}}$ & \textbf{83.0} \\
        \cmidrule(lr){2-3}
        & APT base$^R$ & 80.4 / 80.3 \small{$\pm$0.1} \\
        & APT base$^{R}$ p.e. & \textbf{81.2} \\
        
        \bottomrule
    \end{tabular}
    } % resize box
    \caption{\textsc{Smatch} scores on AMR 1.0, 2.0, and 3.0 test sets.
    % $^*$ uses the cited model with the same RoBERTa embeddings as our model.
    APT is our model.
    $^B$ or $^R$ indicates pre-trained BERT or RoBERTa embeddings, $^G$ use of graph re-categorization,
    $^*$ improved results reported in \citet{lee2020pushing}.
    $^\text{\ding{61}}$ denotes concurrent work based on fine-tuning pre-trained BART large models.
    We report the best/average score $\pm$ standard deviation over 3 seeds.
    p.e. is partial ensemble decoding with 3 seed models.}
    \label{tab:main_amr2.0+1.0}
\end{table}

\begin{table}[!t]
    \centering
    \resizebox{\columnwidth}{!}{%
    \begin{tabular}{cccc}
    \toprule
        \multirow{2}{*}{Model} &
        \multirow{2}{*}{\begin{tabular}{@{}c@{}} Fixed Extra \\ Features \end{tabular}} &
        \multirow{2}{*}{\begin{tabular}{@{}c@{}} Trained \\ Param. \end{tabular}} &
        \multirow{2}{*}{\begin{tabular}{@{}c@{}} \textsc{Smatch} \\ AMR 2.0 \end{tabular}} \\
        \\
    \midrule
        \citet{zhang2019amr}$^{B,G}$ & BERT & 66.1M$^\text{\ding{106}}$  & 76.3 \\ % 174,409,872 % 66,099,600 by removing 'BERT'
        \citet{cai2020amr}$^{B}$ & BERT & 27.1M & 78.7 \\  % 27057335
        \citet{cai2020amr}$^{B,G}$ & BERT & 26.1M & 80.2 \\  % 26054721
        \citet{astudillo2020transition}$^{R}$ & RoBERTa & 21.7M & 80.2 \\ % 21729280
        \citet{xu2020improving} (4M silver) & - & 239.1M & 80.2 \\ %239,120,768
        \citet{bevilacqua2021one}$^{\text{\ding{61}}}$ & - & 411.8M & 83.8 \\  % 411,792,384
    \midrule
        APT small$^{R}$ & RoBERTa & 17.5M  & 81.7 \\ % 17484800
        APT base$^{R}$ & RoBERTa &  21.4M &  81.8 \\ % 21438464
    \midrule
        APT small$^{R}$ p.e. & RoBERTa & 52.5M & 82.5 \\
        APT base$^{R}$ p.e. & RoBERTa & 64.3M  & 82.8 \\

    \bottomrule
    \end{tabular}
    } % resize box
    \caption{Comparison of model parametrization sizes and \textsc{Smatch} scores on AMR 2.0 test set.
    Model sizes of previous works are obtained from their officially released pre-trained models.
    $^\text{\ding{106}}$ is an estimate by removing BERT parameters in the released model, where a BERT base model is trained together which is different from the paper description.
    $^\text{\ding{61}}$ denotes concurrent work based on fine-tuning pre-trained BART large models.
    }
    \label{tab:main_param}
\end{table}

% \begin{table}[!t]
%     \centering
%     \resizebox{\columnwidth}{!}{%
%     \begin{tabular}{ccccc}
%     \toprule
%         \multirow{2}{*}{Model} &
%         \multicolumn{2}{c}{Param. Size} & \multicolumn{2}{c}{\textsc{Smatch} (\%)} \\
%         & Fixed Features & Trained & AMR 1.0 & AMR 2.0  \\
%     \midrule
%         \citet{zhang2019amr}$^{G}$ & BERT & & 70.2 & 76.3 \\
%         \citet{cai2020amr}$^{G}$ & BERT & 26.1M & 75.4 & 80.2 \\  % 26054721
%         \citet{astudillo2020transition} & RoBERTa & & 76.9 & 80.2 \\
%         \citet{xu2020improving} & BERT, 4M silver & & - & 80.2 \\
%         \citet{bevilacqua2021one}$^{\text{\ding{61}}}$ & pre-trained BART & 4.1B & - & 83.8 \\  % 411,792,384
%     \midrule
%         APT small & RoBERTa & 10.9M / 17.5M & 78.2 & 81.7 \\
%         APT base & RoBERTa & 14.9M / 21.4M & 78.5 & 81.8 \\ % 21438464
%     \midrule
%         APT small p.e. & RoBERTa & & 79.7 & 82.5 \\
%         APT base p.e. & RoBERTa & & 79.8 & 82.8 \\

%     \bottomrule
%     \end{tabular}
%     } % resize box
%     \caption{Comparison of model parametrization sizes and \textsc{Smatch} scores on AMR 1.0 and 2.0 test sets.
%     $^\text{\ding{61}}$ denotes concurrent work based on fine-tuning pre-trained BART large models.
%     }
%     \label{tab:main_param}
% \end{table}

%% main results
\paragraph{Parsing Performance}
We compare our action-pointer transition/Transformer (APT) model with existing approaches in Table~\ref{tab:main_amr2.0+1.0}\footnotemark\footnotetext{We exclude \citet{xu2020improving} AMR 1.0 numbers since they report 16833 train sentences, not 10312.}.
We indicate the use of pre-trained BERT or RoBERTa embeddings (from large models) with $^B$ or $^R$, and graph re-categorization with $^G$. Graph re-categorization \cite{lyu2018amr,zhang2019amr, cai2020amr,bevilacqua2021one} removes node senses and groups certain nodes together such as named entities in pre-processing. It reverts these back in post-processing with the help of a name entity recognizer.
We report results over 3 runs for each model with different random seeds. Given that we use fixed pre-trained embeddings, it becomes computationally cheap to build a partial ensemble that uses the average probability of 3 models from different seeds which we denote as p.e.

% put discussion of SPRING into the picture
With the exception of the recent BART-based model \citet{bevilacqua2021one},
we outperform all previously published approaches, both with our small and base models. Our best single-model parsing scores are 81.8 on AMR 2.0 and 78.5 on AMR 1.0, which improves $1.6$ points over the previous best model trained only with gold data.
Our small model only trails the base model by a small margin and 
% that the high performance of previous transition-based parsers in AMR1.0, 
we achieve high performance on small AMR 1.0 dataset,
indicating that our approach benefits from having good inductive bias towards the problem so that the learning is efficient.
% We achieve the new state-of-the-art parsing scores on both the AMR 2.0 and AMR 1.0 datasets. On AMR 2.0, we significantly improve the previous best score of 80.2 \citep{cai2020amr, astudillo2020transition} by 1.5 points on average and our best single model achieves 81.7 \textsc{Smatch}. On AMR 1.0 with more limited training data, we obtain a similar gain of 1.4 on average over the previous best score of 76.9 \citep{astudillo2020transition}, suggesting that our model is also data-efficient.
More remarkably, we even surpass the scores reported in \citet{lee2020pushing} combining various self-learning techniques and utilizing 85K extra sentences for self-annotation (silver data). 
% We believe our approach could also benefit from these techniques for further improvement.
For the most recent AMR 3.0 dataset, we report our results for future reference.

Additionally, the partial ensemble decoding proves to be simple and effective in boosting the model performance, which consistently brings more than $1$ point gain for AMR 1.0 and 2.0. It should be noted that the ensemble decoding is only $20$\% slower than a single model. 

We thus use this ensemble to annotate the $85$K sentence set used in \cite{lee2020pushing}. After removing parses with detached nodes we obtained 70K model-annotated silver data sentences. Adding these for training regularly, we achieve our best score of 83.4 with ensemble on AMR 2.0.

%% model parametrization size
\paragraph{Model Size}
In Table~\ref{tab:main_param}, we compare parameter sizes of recently published models %\footnote{For \citet{zhang2019amr}, we report an estimate by removing BERT parameters in the released model,  where a BERT base model is trained together which is different from the paper description.}
alongside their parsing performances on AMR 2.0.
Similar to our approach, most models use large pre-trained models to extract contextualized embeddings as fixed features, with the exception of \citet{xu2020improving}, which is a seq-to-seq pre-training approach on large amount of data, and \citet{bevilacqua2021one}, which directly fine-tunes a seq-to-seq BART large \citep{lewis2019bart} model.\footnote{Here we focus on trainable parameters for learning efficiency. For deployment the total number of parameters should be considered, where all the models relying on BERT/RoBERTa features would be on the similar level.}
% Although \citet{bevilacqua2021one} holds the highest \textsc{Smatch} score on AMR 2.0, the number of trained parameters is huge compared to other models.\footnote{Here we focus on trainable parameters for learning efficiency. For deployment the total number of parameters should be considered, where all the models relying on BERT/RoBERTa features would be on the similar level.}
Except the large BART model, our APT small (3 layers) has the least number of trained parameters yet already surpasses all the previous models. % using fixed embeddings.
This justifies our method is highly efficient in learning for AMR parsing.
Moreover, with the small parameter size, the partial ensemble is an appealing way to improve parsing quality with minor decoding overhead. Although more performant, direct fine-tuning of pre-trained seq-to-seq models such as BART would require prohibitively large numbers to perform an ensemble.

%% finegrained results
\paragraph{Fine-grained Results}
Table~\ref{tab:finegrained_amr2.0} shows the fine-grained AMR 2.0 evaluation  \citep{damonte2016incremental} of APT and previous models with comparable trainable parameter sizes. Our model achieves the best scores among all sub-tasks except negations and wikification, handled by post-processing on the best performing approach.
% Our model leads previous best sub-task scores consistently by xxx.
We obtain large improvement on edge related sub-tasks including SRL ($\textsc{arg}$ arcs) and Reentrancies, proving the effectiveness of our target-side pointer mechanism.

\begin{table*}
    \centering
    \resizebox{\textwidth}{!}{%
    \begin{tabular}{cccccccccc}
        \toprule
        Model & \textsc{Smatch} & Unlabeled & No WSD & Concepts & Named Ent. & Negations & Wikification & Reentrancies & SRL \\
        \midrule
        \citet{van2017neural}     & 71.0 & 74   & 72   & 82   & 79   & 62   & 65   & 52   & 66 \\
        \citet{groschwitz2018amr}$^G$ & 71.0 & 74   & 72   & 84   & 78   & 57   & 71   & 49   & 64 \\
        \citet{lyu2018amr}$^G$        & 74.4 & 77.1 & 75.5 & 85.9 & 86.0 & 58.4 & 75.7 & 52.3 & 69.8 \\
        \citet{cai2019core} & 73.2 & 77.0 & 74.2 & 84.4 & 82.0 & 62.9 & 73.2 & 55.3 & 66.7 \\
        \citet{naseem2019rewarding}$^B$ & 75.5 & 80 & 76 & 86 & 83 & 67 & 80 & 56 & 72 \\
        \citet{zhang2019amr}$^{B,G}$ & 76.3 & 79.0 & 76.8 & 84.8 & 77.9 & 75.2 & 85.8 & 60.0 & 69.7 \\
        \citet{zhang2019broad}$^{B,G}$ & 77.0 & 80 & 78 & 86 & 79 & 77 & 86 & 61 & 71 \\
        \citet{cai2020amr}$^{B,G}$ & 80.2 & 82.8 & 80.8 & 88.1 & 81.1 & \textbf{78.9} & \textbf{86.3} & 64.6 & 74.2 \\
        \citet{astudillo2020transition}$^*\,$$^{R}$ & 80.2 & 84.2 & 80.7 & 88.1 & 87.5 & 64.5 & 78.8 & 70.3 & 78.2 \\
        \midrule
        APT small$^R$ & 81.7 & 85.4 & 82.2 & \textbf{88.9} & \textbf{88.9} & 67.5 & 78.7 & 70.6 & 80.7 \\
        APT base$^R$ & \textbf{81.8} & \textbf{85.5} & \textbf{82.3} & 88.7 & 88.5 & 69.7 & 78.8 & \textbf{71.1} & \textbf{80.8} \\
        \bottomrule
    \end{tabular}
    } % resize box
    \caption{Fine-grained F1 scores on the AMR 2.0 test set.
    % $^*$ uses the cited model with the same RoBERTa embeddings as our model.
    $^B$/$^R$ and $^G$ marks uses of pre-trained BERT/RoBERTa embeddings and graph re-categorization processing.
    $^*$ We cite improved results reported in \citet{lee2020pushing}.
    We report results with our single best model for fair comparison.}
    \label{tab:finegrained_amr2.0}
\end{table*}

\subsection{Analysis}

\begin{table}[!t]
    \centering
    \resizebox{\columnwidth}{!}{%
    \begin{tabular}{cccc}
    \toprule
        \multicolumn{2}{c}{Model Configuration} & \multicolumn{2}{c}{
        \textsc{Smatch} (\%)} \\
    \midrule
         \begin{tabular}{@{}c@{}} Mono. \\ Alignment \end{tabular} &  \begin{tabular}{@{}c@{}} Graph \\ embedding \end{tabular} & \begin{tabular}{@{}c@{}} AMR \\ 1.0 \end{tabular} & \begin{tabular}{@{}c@{}} AMR \\ 2.0 \end{tabular}  \\
    \midrule
         & & 72.2 \small{$\pm$0.4} & 77.5 \small{$\pm$0.2} \\
         \ding{51} & &  78.0 \small{$\pm$0.1} & 81.5 \small{$\pm$0.1} \\
        %  & \ding{51}  & & 78.3 \small{$\pm$0.1} \\
         \ding{51} & \ding{51}  & 78.3 \small{$\pm$0.1} & 81.7 \small{$\pm$0.1} \\
    \midrule
        \multicolumn{2}{c}{No subspace restriction} & 78.0 \small{$\pm$0.1} & 80.9 \small{$\pm$0.1} \\
        \multicolumn{2}{c}{RoBERTa base embeddings} & 78.0 \small{$\pm$0.1} & 81.3 \small{$\pm$0.1} \\
        \multicolumn{2}{c}{BERT large embeddings} & 77.7 \small{$\pm$0.1} & 81.4 \small{$\pm$0.1} \\
    \bottomrule
    \end{tabular}
    } % resize box
    \caption{Ablation study of model components. The analysis is with our base model size.}
    \label{tab:model_components_2}
\end{table}

\paragraph{Ablation of Model Components}
We evaluate the contribution of different components in our model in Table~\ref{tab:model_components_2}.
%% major components
The top part of the table shows effects of 2 major components that utilize parser state information and the graph structural information in the Transformer decoder. The baseline model is a free Transformer model with pointers (row 1), which is greatly increased by including the  monotonic action-source alignment via hard attention (row 2) on both AMR 1.0 and AMR 2.0 corpus, and combining it with the graph embedding (row 3) gives further improvements of 0.3 and 0.2 for AMR 1.0 and AMR 2.0.
This highlights that injecting hard encoded structural information in the Transformer decoder greatly helps our problem.
% In addition, embedding the intermediate graph structure brings further gains on both of the dataset (row 5-7), and the best results are achieved by having all of the 3 components, which showcased that injecting hard encoded structural information in the Transformer decoder greatly helps our problem.

%% other components
The bottom part of Table~\ref{tab:model_components_2} evaluates the contribution of output space restriction for target and input pre-trained embeddings for source, respectively. Removing the restriction for target output space i.e. the valid actions, hurts the model performance, as the model may not be able to learn the underlying rules that govern the target sequence restrictions. Switching the RoBERTa large embeddings to RoBERTa base or BERT large also hurts the performance (although score drops are only $0.3\sim 0.6$), indicating that the contextual embeddings from large and better pre-trained models better equip the parser to capture semantic relations in the source sentence.

\begin{table}[!t]
    \centering
    \resizebox{\columnwidth}{!}{%
    \begin{tabular}{ccc}
    \toprule
        \multirow{2}{*}{\begin{tabular}{@{}c@{}} Data oracle \\ variation \end{tabular}} &
        \multicolumn{2}{c}{\textsc{Smatch} (\%)} \\
        & Train oracle & Model test \\
    \midrule
        None & 98.9 & 81.7 \small{$\pm$0.1} \\
        \midrule
        No subgraph breakdown & 97.8 &  80.6 \small{$\pm$0.1} \\
        Create farther edges first & 98.9 &  81.4 \small{$\pm$0.2} \\
        Post-order subgraph traversal   & 98.9 &  81.8 \small{$\pm$0.1} \\
    \bottomrule
    \end{tabular}
    } % resize box
    \caption{Results of model performance with different data oracles on AMR 2.0 corpus.}
    \label{tab:transition_oblation}
\end{table}

\paragraph{Effect of Oracle Setup}

As our model directly learns from the oracle actions, we study how the upstream transition system affects the model performance by varying transition setups in Table~\ref{tab:transition_oblation}.
We try three variations of the oracle. In the first setup, we measure the impact of breaking down \textsc{subgraph} action into individual node generation and attachment actions. We do this by using the \textsc{subgraph} for all cases of multi-node alignments. This degrades the parser performance and oracle \textsc{Smatch} considerably, dropping by absolute 1.1 points. This is expected, since \textsc{subgraph} action makes internal nodes of the subgraph unattachable. In the second setup, we vary the order of edge creation actions. We reverse it so that the edges connecting farther nodes are built first. Although this does not affect the oracle score, we observe that the model performance on this oracle drops by 0.3. The reason might be that the easy close-range edge building actions become harder when pushed farther, also making easy decisions first is less prone to error propagation. Finally, we also change the order in which the various nodes connected to a token are created. Instead of generating the nodes from the root downwards, we perform a post-order traversal, where leaves are generated before parents. This also does not affect oracle score, however it gave a minor gain in parser performance.

\begin{figure}[!t]
    \centering
    \includegraphics[width=\columnwidth]{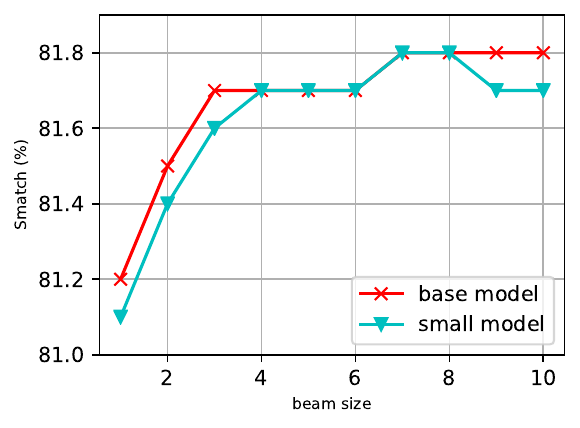}
    \caption{Effect of decoding beam size for \textsc{Smatch}, with our best single models on AMR 2.0 test set.}
    \label{fig:beam}
\end{figure}

\paragraph{Effect of Beam Size}
%% beam size effect; make a plot
Figure~\ref{fig:beam} shows performance for different beam sizes. Ideally, if the model is more certain and accurate in making right predictions at different steps, the decoding performance should be less impacted by beam size. The results show that performance improves with beam size, but the gains saturate at beam size 3. This indicates that a smaller beam size can be considered for application scenarios with time constraints.

\label{sec:results}

\section{Related Work}

With the exception of \citet{astudillo2020transition}, other works introducing stack and buffer information into sequence-to-sequence attention parsers \citep{liu-zhang-2017-encoder,zhang-etal-2017-stack,buys2017robust}, are based on RNNs and do not attain high performances. \citet{liu-zhang-2017-encoder,zhang-etal-2017-stack} tackle dependency parsing and propose modified attention mechanisms while \citet{buys2017robust} predicts semantic graphs jointly with their alignments and compares stack-based with latent and fixed alignments. Compared to the stack-Transformer \citep{astudillo2020transition}, we propose the use of an action pointing mechanism to decouple word and node representation, remove the need for stack and buffer and model graph structure on the decoder side. We show that these improvements yield superior performance while exploiting the same inductive biases  with little train data or small models.

\citet{vilares2018transition} proposed an \textsc{amr-convington} system for unrestricted non-projective AMR parsing, comparing the current word with all previous words for arc attachment as we propose. However, their comparison is done with sequential actions whereas we use an efficient pointer mechanism to parallelize the process.

Regarding the use of pointer mechanisms for arc attachment, \citet{ma2018stack} proposed the stack-pointer network to build partial graph representations, and \citet{fernandez2020transition} adopted pointers along with the left-to-right scan of the sentence, greatly improving the efficiency. Compared with these works, we tackle a more general text-to-graph problem, where nodes are only loosely related to words, by utilizing the action-pointer mechanism. Our method is also able to build up to depth $M$ graph representations with $M$ decoding layers.

While not explicitly stated, graph-based approaches \citep{zhang2019amr, cai2020amr} generate edges with a pointing mechanism, either with a deep biaffine classifier \citep{dozat2018simpler} or with attention \citep{vaswani2017attention}.
% Note that \citet{zhang2019amr} also used a target side pointer, but their usage is dealing with duplication of nodes since they transferred graphs to trees, which is quite different from our design.
They also model inductive biases indirectly through graph re-categorization, detailed in Section~\ref{results:main}, which requires a name entity recognition system at test time. Re-categorization was proposed in \citet{lyu2018amr}, which reformulated alignments as a differentiable permutation problem, interpretable as another form of inductive bias.

% \citet{gildea2018cache, peng2018sequence} used a cache transition system to handle reentrant edges with a similar idea by comparing nodes stored in a cache, but they process node generation and edge construction with two separate stages, and actions work on predicted nodes instead on the original sequence.

%One major advantage of the transition-based approaches over the graph-based approaches is that they can encode rich local context information from the parser states, which proves very important for neural AMR parsing given the limited amount of annotated data. The parser states, such as stack and buffer, are encoded in the neural models in different forms, such as with LSTMs \citep{dyer2015transition, ballesteros2017amr, peng2018sequence} and Transformers \citep{astudillo2020transition}. Our handling of the action-source hard alignment can also be viewed as one such example. 

%Interestingly, recent work on utilizing large amount of external data and pre-trained models for AMR parsing \citep{xu2020improving} shows strong results without making use of any specialized information, which could also be an interesting future direction to explore with our system.

Finally, augmenting seq-to-seq models with graph structures has been explored in various NLP areas,
including machine translation \citep{hashimoto2017neural, moussallem2019augmenting}, text classification \citep{lu2020vgcn}, AMR to text generation \citep{zhu2019modeling}, etc.
% For example, \citet{hashimoto2017neural, moussallem2019augmenting} embed graph structures for neural machine translation, with either a latent parsing graph or an external knowledge graph, \citet{lu2020vgcn} combines the graph neural network with BERT for text classification, and \citet{zhu2019modeling} embeds the graph structure in Transformer for better AMR to text generations.
Most of these works model graph structure in the encoder since the complete source sentence and graph are known. We embed a dynamic graph in the Transformer decoder during parsing. This is similar to broad graph generation approaches \citep{li2018learning} relying on graph neural networks \citep{li2019hierarchy}, but our approach is much more efficient as we do not require heavy re-computation of node representations.

\section{Conclusion}

% We propose the Action-Pointer Transformer model which combines hard-attention with a target-side action  pointer mechanism to yield a more expressive transition-parser that retains strong inductive biases. We achieve new state-of-the art results for AMR parsing while showing also good performance for smaller models or limited training data.

We present an Action-Pointer mechanism that can naturally handle the generation of arbitrary graph constructs, including re-entrancies and multiple nodes per token.
Our structural modeling with incremental encoding of parser and graph states based on a single Transformer architecture proves to be highly effective, obtaining the best results on all AMR corpora among models with similar learnable parameter sizes.
An interesting future exploration is on combining our system with large pre-trained models such as BART, %similar to \citep{bevilacqua2021one}.
as directly fine-tuning on the latter shows great potential in boosting the performance \citep{bevilacqua2021one}.
Although we focus on AMR graphs in this work, our system can essentially be adopted to any task generating graphs from texts where copy mechanisms or hard-attention plays a central role.

% future work: similar approach could be used for other conditional graph/formal language generation problems, e.g. programming language?, as long as we can find the appropriate description of how the graph can be generated via some sequential operations 

% \section{Broader Impact Statement}
% Ethical consideration? Do we need? (not in 8 pages)
% ethical considerations; not counted in 8 pages
%\input{figure1}

\bibliography{anthology,custom}
\bibliographystyle{acl_natbib}

\newpage

\appendix

% \section{Example Appendix}
% \label{sec:appendix}

% This is an appendix.

% \section{Valid Actions}

\section{A More Detailed Example of Action-Pointer Transitions}

\begin{figure*}[t]
    \centering
    \includegraphics[width=\textwidth]{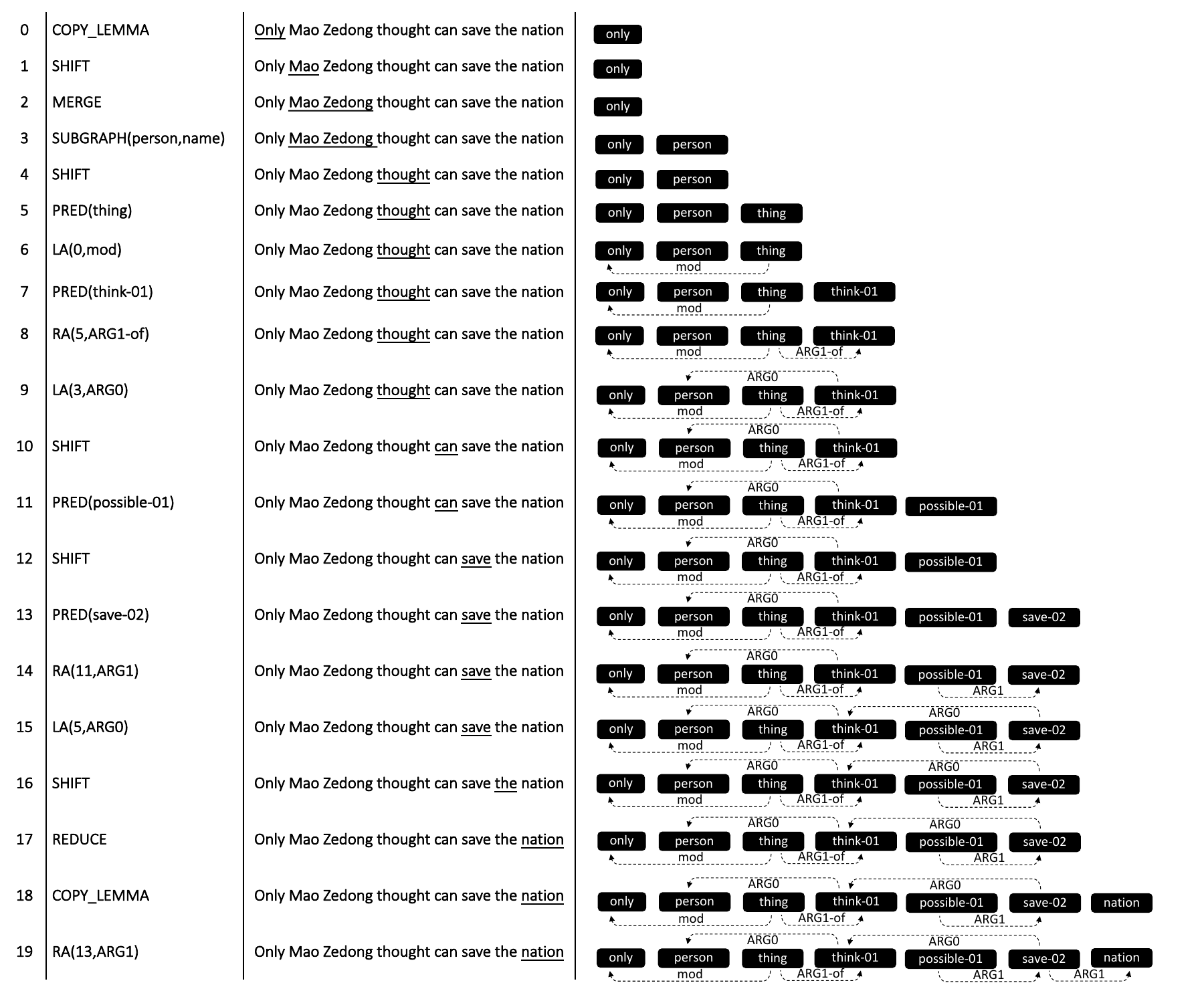}
    \caption{Step-by-step actions based on our action-pointer transition system. We illustrate the use of \textsc{merge} and \textsc{subgraph} with the named entity of a person's name in this example. The source cursor after the action is applied is pointing at words underlined, and the partially built graph is shown in the right-most column.}
    \label{fig:maozedong}
\end{figure*}

We present a step-by-step walk-through of our actions on a less trivial example for generating the AMR in Figure~\ref{fig:maozedong}. The sentence contains a named entity which also demonstrates the $\merge$ and \textsc{subgraph} usage of our transition system.

\section{Action-Pointer Oracle}\label{a:oracle}

For a given sentence, at every oracle step, apply the actions in the order listed below. Continue until the source cursor moves past the last token.

\begin{enumerate}
\item \textbf{If} cursor is on a non-final token of a span aligned to a node, keep moving cursor (and collecting tokens) with $\merge$ till it reaches the final token of the span. 
\item \textbf{If} there is a matching pattern for current token(s) in $\subgraph{}$ action dictionary:
\begin{itemize}
\item Apply matching $\subgraph{}$ action.
\item Generate edges between the root of the sub-graph and past nodes through $\la{}$, $\ra{}$. Generate closer edges first.     
\end{itemize}
\textbf{Otherwise}, for all nodes aligned to the current token, in top-down order:
\begin{itemize}
\item Generate node through $\cp$ (lemma or first sense), and if not possible then through $\pred{}$. 
\item Generate edges between the last nodes and past nodes through $\la{}$, $\ra{}$. Generate closer edges first.   
\end{itemize}
\item \textbf{If} no action performed at step 2, move cursor with $\reduce$ \textbf{otherwise}, move cursor with $\shift$. 
\end{enumerate}

\section{Action-Pointer Decoding}\label{b:decoding}

We outline the decoding algorithm for our model in Algorithm \ref{algo:decoding}, to combine the actions with pointers, as well as taking in parsing states and graph structures for the model during the decoding steps. Detailed beam search process is ignored. Although tacking the specific problem of AMR graph generation with pointers, our constraint decoding process is a modified beam search algorithm with different components and step-wise controls, among others \citep{vijayakumar2016diverse, zhou2019simple}.

\begin{algorithm}[h]
\SetAlgoLined
\KwIn{Initial token $a_0=$</s>, beam size $k$, max step $T_{max}$, action dictionary $D$ without pointers, model $M$ that outputs both distribution over $D$ and the pointer distribution from self-attention}
\KwOut{Decoded results $y_1=(a_1, p_1), y_2=(a_2, p_2), \cdots, y_T=(a_T, p_T)$}
% \KwResult{Write here the result }
 initialization: step $t=1$, $k$ state machines \\
 \While{$t<=T_{max}$}{
  1) Get the valid action dictionary $D_t\subset D$, previous node action positions $N_t\subset \{0, 1, 2, \ldots, t\}$, current token cursor $c_t$, and current graph $G_t$ (all from the corresponding state machines); \\
  2) Input prefix $a_0, a_1, \cdots, a_{t-1}$ and $D_t, c_t, G_t$ into model, get output distribution $\prob{a_t | \by_{<t}}$, and the self-attention distribution $Q(p)$ from pointer head with $p$ over $\{0, 1, \ldots, t\}$; \\
  3) Take the most likely valid pointer value, with $p^*=\argmax_{p\in N_t}Q(p)$, and its score $q^*=\max_{p \in N_t}Q(p)$;\\
%   For each possible action $a$ from $D$:\\
%   \eIf{$a$ is an edge action}{
%   combine the action probability with pointer probability $\prob{y_t}=\prob{a_t | \by_{<t}}\cdot q^*$, with $y_t=(a_t, p^*)$
%   }{
%   set $\prob{y_t}=\prob{a_t | \by_{<t}}$, with $y_t=(a_t, null)$
%   }
  \For{each possible action $a$ from $D$} {
\eIf{$a$ is an edge action}{
  combine the action probability with pointer probability $\prob{y_t}=\prob{a_t | \by_{<t}}\cdot q^*$, with $y_t=(a, p^*)$
   }{
   set $\prob{y_t}=\prob{a_t | \by_{<t}}$, with $y_t=(a, null)$
  }
}
 Do beam search with $\prob{y_t}$ over $y_t$ to get $k$ decoded results;\\
 Apply the corresponding actions with the $k$ state machines to update parser states and partial graphs for each beam candidate.
 }
 \caption{Constrained beam search for action-pointer decoding}\label{algo:decoding}
\end{algorithm}

\section{Number of Parameters}

Our model is a single Transformer \citep{vaswani2017attention} model. The pointer distribution, action-source alignment encoding from parsing state, and structural graph embedding are all contained in certain attention layers and heads, without introducing any extra parameters on original Transformer.
We fix our model size and all the embedding size to be 256, and the feedforward hidden size in Transformer as 512. And they are the same for our base model with 6 layers and 4 heads and our small model with 3 layers and 4 heads, both for encoder and decoder.

% The source side dictionary is built from English sentences from training data. The dictionary size is 33256 for AMR 2.0 and 16824 for AMR 1.0. 44536 for AMR3.0
We use pre-trained RoBERTa embeddings for the source token embeddings. The embeddings are extracted in pre-processing and fixed. The RoBERTa model parameters are fixed and not trained with our model.
We have a projection layer to project the RoBERTa embedding size 1024/768 to our model size 256.

The target side dictionary is built from all the oracle actions without pointers on training data. The dictionary size for AMR 1.0 is 4640, for AMR 2.0 is 9288, and for AMR 3.0 is 11680. % 12008.
We build the target action embeddings along with the model for the action prediction on top of Transformer decoder. The dictionary embedding size is fixed at 256.

Overall, the total number of parameters for our 6 layer base model is 14,852,096 on AMR 1.0, 21,438,464 on AMR 2.0, and 25,550,848 %25,718,784
on AMR 3.0 (difference is in target dictionary embedding size).
The total number of parameter for our 3 layer small model is 10,898,432 for AMR 1.0 and 17,484,800 on AMR 2.0 (difference is in target dictionary embedding size).

% \section{AMR Pre- and Post- Processing}

% re-catogrization in details in other works?

\end{document}